\documentclass[runningheads]{llncs}

\usepackage{times}
\usepackage{epsfig}
\usepackage{graphicx}
\usepackage{amsmath}
\usepackage{amssymb}
\usepackage{multirow}
\usepackage{amsfonts}
\usepackage{dsfont}
\usepackage{indentfirst}

\usepackage{hyperref}

\newcommand{\eg}{\emph{e.g.}\ }
\newcommand{\etal}{\emph{et al.}\ }

\begin{document}

\title{IterDet: Iterative Scheme for Object Detection in Crowded Environments}

\author{
    Danila Rukhovich \and
    Konstantin Sofiiuk \and
    Danil Galeev \and
    Olga Barinova \and
    Anton Konushin
}
\authorrunning{D. Rukhovich \etal}
%
\institute{Samsung AI Center\\
\email{\{d.rukhovich,k.sofiiuk,d.galeev,o.barinova,a.konushin\}@samsung.com}}

\maketitle

\begin{abstract}
Deep learning-based detectors tend to produce duplicate detections of the same objects. After that, the detections are filtered via a non-maximum suppression algorithm (NMS) so that there remains only one bounding box per object. This simple greedy scheme is sufficient for isolated objects. However, it often fails in crowded environments since boxes for different objects should be preserved and duplicate detections should be suppressed at the same time. In this work, we propose to obtain predictions following \emph{iterative scheme} called IterDet. At each iteration, a new subset of objects is detected. Detected boxes from all the previous iterations are considered at the current iteration to ensure that the same object would not be detected twice. This iterative scheme can be applied to both one-stage and two-stage deep learning-based detectors with minor modifications. Through extensive evaluation on 4 diverse datasets with two different baseline detectors, we prove our iterative scheme to achieve significant improvement over the baseline. On CrowdHuman and WiderPerson datasets, we obtain state-of-the-art results. The source code and the trained models are available at \url{https://github.com/saic-vul/iterdet}.
\end{abstract}

\section{Introduction}
\label{sec:intro}
In recent years, deep learning-based methods of object detection have significantly evolved and achieved solid improvements in terms of speed and accuracy \cite{liu2016ssd,girshick2015fast,ren2015faster,lin2017focal,tian2019fcos}.

All deep learning-based detectors densely sample and independently evaluate candidate object locations. Accordingly, for a single object, they yield multiple similar boxes of varying confidence. This redundant set of detected boxes is then filtered via non-maximum suppression (NMS) or similar techniques to produce exactly one bounding box per object. This \emph{greedy scheme} is sufficient if instances of the same class do not overlap in the image. 

However, this is not always the case. Another possible scenario for object detection is so-called crowded environments that contain multiple overlapping objects of the same class (\eg people in the street or bacteria in microscopy images). Crowded environments provide a challenging task for object detectors due to several reasons. First, it is extremely difficult to distinguish whether two candidate boxes belong to the same object or correspond to two overlapping objects. Second, weak visual cues of heavily occluded instances can hardly provide sufficient information for accurate object detection.

In several works, this problem has been addressed with various modifications of the NMS algorithm \cite{rothe2014non,bodla2017soft,hosang2017learning,tychsen2018improving,liu2019adaptive,huang2020nms}. By NMS, both duplicate detections of the same object should be removed and the hard-to-detect occluded objects should be kept at the same time. Therefore, there is a natural trade-off between precision and recall that imposes severe restrictions on all these approaches. 

\begin{figure*}[t]
\centering
\begin{tabular}{cc}
    \includegraphics[width=0.47\linewidth]{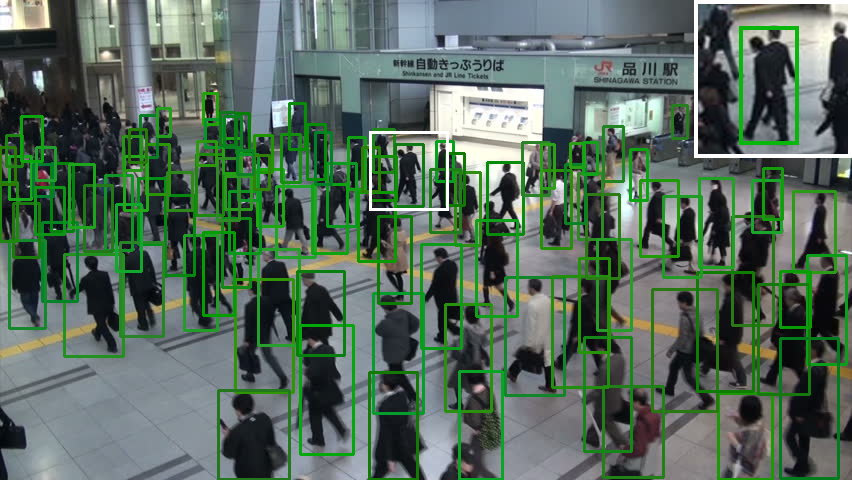} &
    \includegraphics[width=0.47\linewidth]{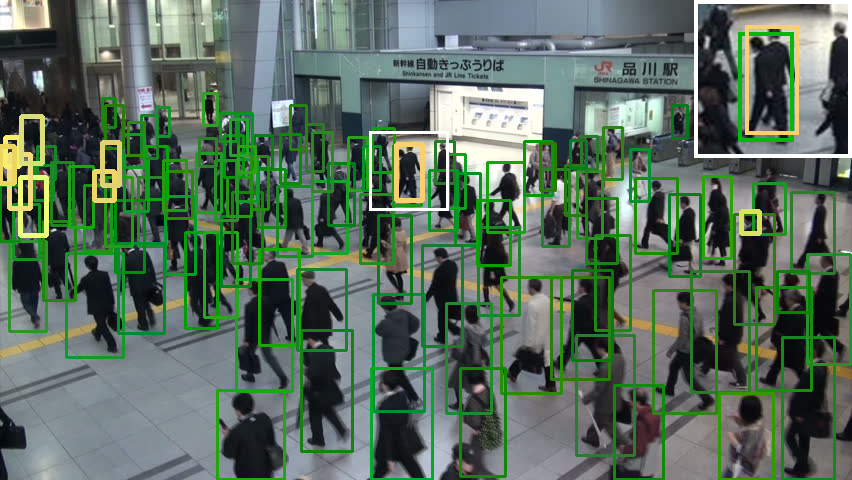} \\
    {\small \emph{baseline} recall: 78.8, AP: 76.81} & {\small \emph{IterDet 1 iter.} recall: 75.9, AP: 74.28} \\
    & {\small \emph{IterDet 2 iter.} recall: \textbf{82.5}, AP: \textbf{79.59}}
\end{tabular}
\caption{The results of original Faster RCNN (left) and the proposed IterDet based on Faster RCNN (right) for the same image from CrowdHuman \emph{test} set with \emph{visible-body} annotations. The boxes found on the first and second iteration are marked in green and yellow, respectively. The metrics for baseline and IterDet after the first and the second iterations are listed below the images.}
\label{fig:teaser}
\end{figure*}

In this work, we describe a novel \emph{iterative scheme} (IterDet) for object detection. Rather than detecting all objects in the image simultaneously, we propose to yield detections iteratively. At each iteration, a new subset of objects is detected. Object boxes that have been already found at the previous iterations are passed to the neural network at the current iteration, so duplicates can be avoided. The proposed iterative scheme can be applied to any one-stage or two-stage object detection method with only minor modifications.

Figure \ref{fig:teaser} demonstrates the results of IterDet for Faster RCNN \cite{ren2015faster} on a test image from CrowdHuman dataset \cite{shao2018crowdhuman}. True positive boxes with scores above 0.1 are visualized, and false positives are omitted for visual clarity. At the second iteration, 9 additional objects (shown in yellow) out of 137 are detected, overtaking the baseline Faster RCNN by 5 true positives and 2.7\% of average precision (AP). In the top-right corner of the images, there is an example of two strongly overlapping objects detected with IterDet yet missed by the baseline detector.

Recently, there have been introduced several neural architectures that handle image context thus being more suitable for crowded environments \cite{hu2018relation,goldman2019precise,ge2020ps}. For instance, \cite{stewart2016end} proposed to use a special Hungarian loss function to train a convolutional-recurrent model that yields strictly one detection per iteration. In comparison, our approach is more computationally efficient. Moreover, instead of storing information about previously detected objects via LSTM, we explicitly pass it to the network in a form of object masks. Our approach guarantees that no previously detected bounding boxes are accidentally forgotten. Furthermore, compared to \cite{stewart2016end} it allows incorporating the history of detections into deeper layers of a neural network. 

In PS-RCNN \cite{ge2020ps}, objects are also detected iteratively: simple objects are supposed to be found on the first iteration while the second iteration is performed to explore more difficult cases. This iterative approach can be applied only for RCNN-based detectors. At the same time, our approach can be easily integrated into state-of-the-art object detection methods.

We perform extensive experiments with both one-stage (RetinaNet \cite{lin2017focal}) and two-stage (Faster RCNN \cite{ren2015faster}) object detectors on four challenging datasets (AdaptIS ToyV1 and ToyV2 \cite{sofiiuk2019adaptis}, CrowdHuman \cite{shao2018crowdhuman}, and WiderPerson \cite{zhang2019widerperson}). To prove our ideas, we evaluate IterDet against baseline models and compare the obtained results with the results reported by competitors. On all datasets, IterDet outperforms baseline models and sets new state-of-the-art on CrowdHuman and WiderPerson datasets.

\section{Related work}
\label{sec:related}

\textbf{Standard methods for object detection.}
Deep learning-based object detectors can be classified as two-stage and one-stage detectors. Two-stage detectors are based on proposal-driven mechanism \cite{girshick2015fast,ren2015faster}. They consist of two subnetworks: the first one outputs a sparse set of candidate object locations and the second one classifies these object locations into one of the foreground classes or a background. 

One-stage methods are applied over a regular, dense sampling of object locations, scales, and aspect ratios \cite{liu2016ssd,lin2017focal}. Being much faster on inference than their two-stage counterparts, recent one-stage methods achieve comparable accuracy on some datasets. Moreover, anchor-free one-stage methods \cite{tian2019fcos} are more agile and less limited compared to their predecessors. However, two-stage methods still demonstrate state-of-the-art accuracy on challenging datasets.

Overall, all detectors have certain pros and cons and are applicable under certain conditions. To cover all possible scenarios, we design our iterative scheme so it can be combined with both one-stage and two-stage object detectors.

For deep learning-based methods, the detection problem is formulated as a classification task. Namely, class probabilities are estimated independently for each location for multiple candidate locations across an image. Differently, in our iterative scheme, the history of detections from the previous iterations is passed to the detector at the following iterations, providing the context for resolving ambiguities.

\textbf{Modifications of NMS algorithm.}
The standard NMS algorithm greedily selects detections with a higher score and removes the less confident neighbors. Thus, a wide suppression parameter improves the precision and the narrow suppression improves the recall. Consequently, crowded environments are the most challenging case for NMS since both wide and narrow suppression lead to errors. To address this, numerous modifications of the NMS algorithm have been proposed in the literature. Rothe \etal \cite{rothe2014non} formulated NMS as a clustering problem. Hosang \etal \cite{hosang2017learning} suggested decreasing the confidence of detections that cover the already detected objects. In soft NMS \cite{bodla2017soft}, scores for object proposals depend on their overlap with a target object. In adaptive NMS \cite{liu2019adaptive}, parameters of NMS are chosen according to the density of the objects estimated via an extra branch. Most recent R$^2$NMS \cite{huang2020nms} simultaneously predicts the full and visible boxes of an object. 

Differently from all the listed methods, our proposed scheme is iterative that gives more freedom and flexibility. More specifically, we might miss the more difficult objects at the first iteration, since these objects can be detected later on. Accordingly, we do not need to assure high recall at each iteration as we can set wider suppression parameters to favor precision.

\textbf{Neural architectures for crowded environments.}
Several neural architectures for object detection in crowded environments have been described in the literature. Stewart \etal \cite{stewart2016end} used a Hungarian loss function to train an LSTM-based model that yields a sequence of detections. LSTM was also used in \cite{gong2019improving} for iterative proposal refinement in RPN-based detectors. However, performing the NMS step after all iterations negates all the benefits in case of crowded environments. Hu \etal \cite{hu2018relation} proposed an object relation module that processes a set of objects based on their visual appearance and geometry. Ge \etal \cite{ge2020ps} introduced a modification of two-stage detectors called PS-RCNN. First, it detects non-occluded objects with RCNN and then suppresses the detected instances with object-shaped masks. At the second step, another RCNN detects occluded objects.

Compared to the aforementioned methods, our iterative scheme does not imply changing neural architecture, therefore it is much easier to implement.

\section{Proposed method}
\label{sec:method}

\begin{figure*}[ht]
    \centering
        \includegraphics[width=0.96\linewidth]{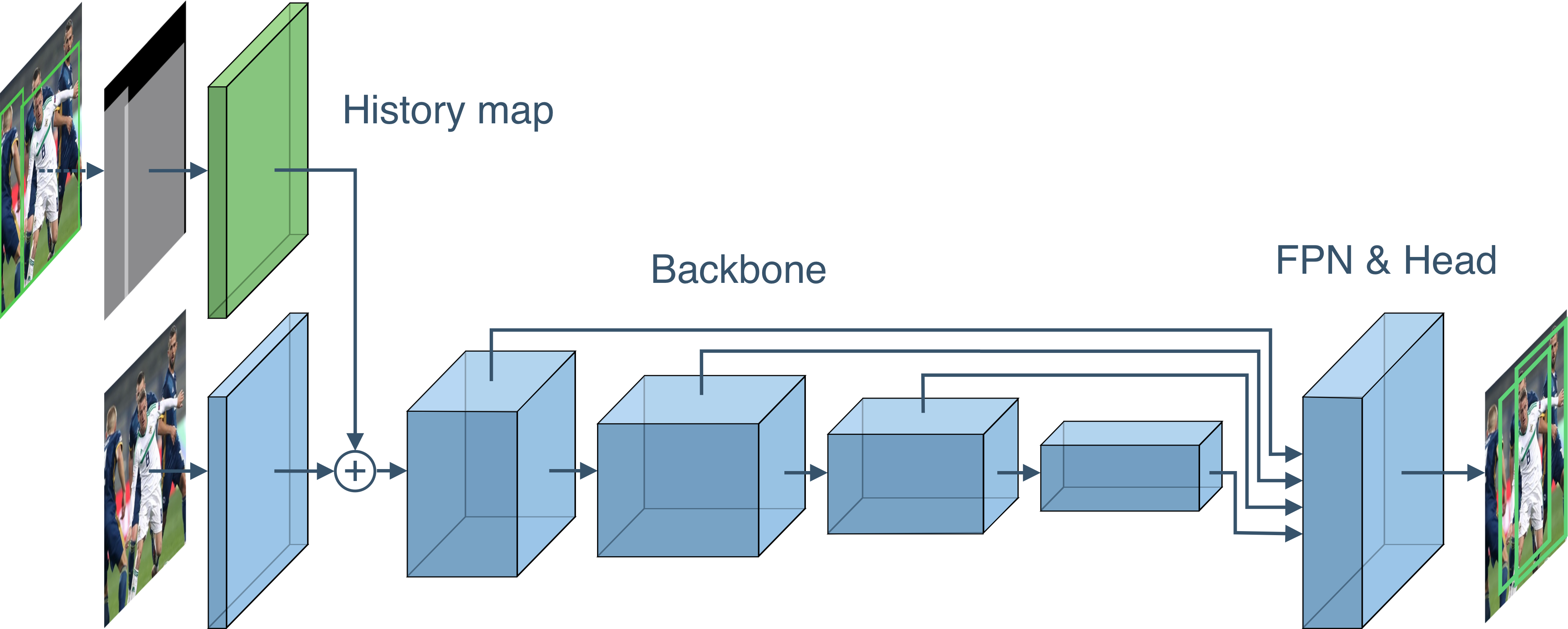}
    \caption{Proposed iterative scheme. The unchanged meta-architecture of an arbitrary detector is marked with blue. The single convolution layer for the history map is marked green. Out of the 4 overlapping objects in the image, 2 are in the history, where they were either randomly sampled at the training step, or detected during previous iterations of the inference. The remaining 2 are predicted by the detector.}
    \label{fig:scheme}
\end{figure*}

The proposed iterative scheme is shown in Figure \ref{fig:scheme}. First, we introduce notation and describe the inference process. Then, we describe the modified training procedure.

\textbf{Inference process.} A typical object detector $D$ is an algorithm that maps image $I \in \mathbb{R}^{w \times h \times 3}$ to a set of bounding boxes $B=\{(x_k, y_k, w_k, h_k)\}_{k=1}^{n}$. Each box is represented by the coordinates of its top left corner $(x, y)$, width $w$ and height $h$. For a given set of boxes $B$, we define a history image $H \in \mathbb{Z}^{w \times h}$ of the same size as an input image. In $H \in \mathbb{Z}^{w \times h}$, each pixel contains the number of already detected boxes that cover that pixel:
\begin{equation}
H_{xy}=\sum_{k=1}^{|B|}{\mathds{1}_{\ x_k \le x \le x_k + w_k,\ y_k \le y \le y_k + h_k}}
\end{equation}
Figure \ref{fig:scheme} shows an example of the history, where its values are color-coded. We can make a detector $D'$ history-aware if we pass the history $H$ along with the image $I$ as its inputs.

Let us now introduce the iterative scheme $IterDet(D')$, that, given an image $I$, produces a set of bounding boxes $B$ in an iterative manner. At the first iteration $t=1$ history $H_1$ is empty and $D'$ maps an image $I$ and $H_1$ to a set of bounding boxes $B_1$. Second, $B_1$ is mapped to history $H_2$. $H_2$ is then mapped to $B_2$ by $D'$ at iteration $t=2$. This process stops when the limit of iterations is reached or when $|B_m|=0$ at some iteration $m$. The final prediction of $IterDet(D')$ is $B=\bigcup_{t=1}^{m} B_{t}$, where $m$ denotes the total number of iterations.

To implement the described scheme, two modifications should be implied: 1) an arbitrary detector $D$ should be altered to become a history-aware detector $D'$ and 2) $D'$ should be forced to predict different sets of objects $B_t$ on each iteration $t$. Below, we explain these alterations in detail.

\textbf{Architecture of a history-aware detector.} State-of-the-art deep learning-based object detection pipelines start with passing an image to an already pre-trained backbone, \eg ResNet \cite{he2015deep}, VGG \cite{simonyan2014very}, etc. to obtain multi-level image features. These features are then fed into trainable feature extractors, \eg Region Proposal Network, Feature Pyramid Network, etc. Their outputs are further passed to the head module predicting bounding boxes. Finally, non-maximum suppression is applied. We aim to introduce minimal changes to the original network architectures and incorporate history in the deepest layers of the network.

The proposed architecture of the history-aware detector is simple yet efficient. The history is processed via one convolution layer which output sums up with the output of the first convolution layer of the backbone. This scheme can be applied to any backbone without hyperparameter tuning. For ResNet-like backbone, the image is passed through a convolution layer with 64 filters of size 7 and stride 2, Batch Normalization layer, and ReLU activation layer. We follow the design choices of ResNet and use a convolution layer with 64 filters of size 3 and stride 2.

\textbf{Training procedure.} During training, we randomly split the set of ground truth bounding boxes $\hat{B}$ into two subsets $B_{old}$ and $B_{new}$ such that $B_{old} \cup B_{new}=\hat{B}$ and $B_{old} \cap B_{new}=\emptyset$. We map $B_{old}$ to a history $H$ and force $D'$ to predict the bounding boxes $B_{new}$ that are missing in history. Thus, we optimize the losses of $D'$ by back-propagation of the error between the predicted boxes $B$ and target boxes $B_{new}$. We do not describe losses since we do not modify this part of baseline detectors. On the one hand, this method of training forces the model to exploit the history and predict only new objects at each iteration of inference. On the other hand, it provides an additional source of augmentations by sampling different combinations of $B_{old}$ and $B_{new}$.

Several iterative methods predict only one object per iteration \cite{barinova2012detection,stewart2016end}. Our iterative scheme is also able to predict one object per iteration \eg by selecting the most confident detection. However, in practice, such an approach would be inefficient since inference time is proportional to the number of objects in the image. Our experiments in Section \ref{sec:experiments} demonstrate that performing two iterations is enough to achieve the best accuracy. With increasing the number of iterations, the recall improves but the precision degrades, worsening mMR and AP metrics. 

\section{Experiments}
\label{sec:experiments}

\subsection{Datasets and implementation details}

We evaluate our proposed iterative scheme on four datasets containing images of various crowded environments: AdaptIS ToyV1 and ToyV2 \cite{sofiiuk2019adaptis}, CrowdHuman \cite{shao2018crowdhuman} and WiderPerson \cite{zhang2019widerperson}.

\begin{table*}[t]
    \centering
    \begin{tabular}{c|cccc}
        \hline
        & Toy V1 & Toy V2 & CrowdHuman & WiderPerson \\ \hline \hline
        object/image & 14.88 & 31.25 & 22.64  & 29.51 \\ \hline
        pair/image & & & & \\
        IoU $>$ 0.3 & 3.67 & 7.12 & 9.02 & 9.21\\
        IoU $>$ 0.4 & 1.95 & 3.22 & 4.89 & 4.78\\
        IoU $>$ 0.5 & 0.95 & 1.25 & 2.40 & 2.15\\
        IoU $>$ 0.6 & 0.38 & 0.45 & 1.01 & 0.81\\ \hline
    \end{tabular}
    \caption{Average number of objects and pair-wise overlap between two instances on the four datasets used in our experiments.}
    \label{tab:datasets}
\end{table*}

\textbf{AdaptIS.} AdaptIS Toy V1 and Toy V2 are two synthetic datasets originally used for instance segmentation task \cite{sofiiuk2019adaptis}. Each image contains about 30 objects on average, with many of those severely overlapping. The datasets statistics are summarized in Table \ref{tab:datasets}. For Toy V1, training and validation splits contain 10000 and 2000 images of size $96 \times 96$ pixels respectively. Toy V2 is split into training, validation, and test subsets with 25000, 1000, and 1000 images of size $128 \times 128$ pixels respectively. For both Toy datasets, we chose AP as the main metric, and also provide recall values for consistency. We do not report the mMR metric: if the average number of false positives per image is less than 1 it turns zero, thus being not representative.

\textbf{CrowdHuman.} The recently introduced CrowdHuman dataset has the largest number of persons per image and the largest number of pairs of intersecting bounding boxes among all datasets for human detection, according to \cite{shao2018crowdhuman}. It contains 15000, 4370, and 5000 images for training, validation, and testing, respectively. There are 23 people presenting on an average image, each annotated with 3 boxes: \emph{full-body}, \emph{visible-body} and \emph{head} box. The most challenging and most frequently used in other works is full-body annotation, where the boxes not only overlap more strongly but also go beyond the edges of the image. We also conduct experiments on visible-body annotation, training models on the training part of the data, and benchmarking on the validation subset.

\cite{shao2018crowdhuman} also reports metrics for one-stage RetinaNet detector and the two-stage Faster RCNN detector, both using ResNet-50 as a backbone. The mMR metric is proposed as the major metric to evaluate detection quality. This metric is calculated as the logarithm of missing rate averaged over 9 points ranging from $10^{-2}$ to $10^0$ false positives per image. Besides, recall and average precision (AP) are reported.

\textbf{WiderPerson.} WiderPerson \cite{zhang2019widerperson} is another human detection dataset collected from various sources. There are 8000, 1000, and 4382 images in train, validation, and test subsets. It contains annotations for 5 classes: pedestrians, riders, partially visible persons, crowd, and ignored regions. Following \cite{ge2020ps}, we merge the last four types into one category for both training and testing.

\textbf{Implementation details.} Our implementation of the proposed IterDet and all baseline models is based on the MMDetection framework \cite{mmdetection}. This framework is built on top of the PyTorch library \cite{paszke2019pytorch} and contains implementations of numerous object detection models. For our experiments, we use RetinaNet and Faster RCNN based on ResNet-50 with default parameters. We use 8 GPUs with 2 images per each. 
The minor modifications are described below. First, we add a Batch Normalization layer after each convolution layer to the FPN of both detectors, which slightly improves performance. Secondly, we do not freeze the first block of ResNet as we add history together with the trainable convolution layer before this block. 
To simplify the hyperparameter tuning in IterDet experiments, we use Adam optimizer with an initial learning rate of 0.0001. For the baseline experiments, we use SGD optimizer with momentum 0.9, weight decay parameter 0.0001, and the initial learning rate 0.02. The training process finishes at the end of the 24th epoch, and the learning rate is multiplied by 0.1 after 16th and 22nd epochs.

\textbf{Dataset-specific hyperparameters.} To be consistent with the CrowdHuman benchmark on inference, the input image is re-scaled such that its shortest edge is 800 pixels, and the longest side is not beyond 1400 pixels. We do not use test-time augmentations. During training, we apply horizontal flips and zooming from $75\%$ to $125\%$. When training on CrowdHuman, we use information about ignored regions when sampling negative examples. For experiments with \emph{full-body} annotations on CrowdHuman, we follow \cite{shao2018crowdhuman,ge2020ps,liu2019adaptive} using $[1.0,\ 1.5,\ 2.0,\ 2.5,\ 3.0]$ anchor ratios and no clipping proposals. For AdaptIS Toy V1 and Toy V2 datasets, we upscale images to $384 \times 384$ pixels as it have been proposed in an original paper \cite{sofiiuk2019adaptis}. The experimental protocol for the WiderPerson dataset is identical to the CrowdHuman dataset.

\subsection{Results and discussion}

\textbf{Results on AdaptIS datasets.} Table \ref{tab:toy} summarizes IterDet and baseline metrics on AdaptIS Toy V1 and Toy V2 datasets. For both datasets and detectors, IterDet substantially increases AP. For Faster RCNN, this increase expands 4\% bringing the final AP up to 99\%.

\textbf{Results on CrowdHuman.} Results on full-body and visible-body annotations of the CrowdHuman dataset are presented in Tables \ref{tab:crowd_human_full} and \ref{tab:crowd_human_visible} respectively. We compare the proposed IterDet scheme to the methods that do not use additional data or annotations during training. According to Table \ref{tab:crowd_human_full}, we achieve a significant improvement in terms of all metrics for the most challenging \emph{full-body} annotation. More specifically, IterDet improves recall by more than 5.5\%, AP by 3.1\% and mMR by 1.0\% w.r.t. baseline. These results remain solid even when compared to the previous state-of-the-art approaches such as Adaptive NMS and PS-RCNN. In terms of mMR, IterDet outperforms all existing methods in all four scenarios: single- and two-stage detectors, visible- and full-body annotations. For the RetinaNet detector, the quality gap exceeds 6\% for both types of annotations. Notice, that such an improvement of mMR value is achieved even after the first iteration. We attribute this to the regularization provided by history-aware training. Despite a slight degradation of mMR with an increasing number of iterations, the growth of AP always remains significant. For RetinaNet, we outperform the competitors by 3.9\% AP for both types of annotations.

\begin{table*}[h!]
    \centering
    \begin{tabular}{c|c|cc|cc}
        \hline
        \multirow[b]{2}{*}{Method} & \multirow[b]{2}{*}{Detector} & \multicolumn{2}{c|}{Toy V1} & \multicolumn{2}{c}{Toy V2} \\ \cline{3-6}
        && Recall & AP & Recall & AP \\ \hline \hline
        Baseline & \multirow{3}{*}{RetinaNet} & 95.46 & 94.46 & 96.27 &	95.62 \\ \cline{1-1} \cline{3-6}
        IterDet, 1 iter. && 95.21 & 95.31 & 96.27 & 94.17 \\
        IterDet, 2 iter. && \textbf{99.56} & \textbf{97.71} & \textbf{99.35} & \textbf{97.27} \\ \hline \hline
        Baseline & \multirow{3}{*}{Faster RCNN} & 94.05 & 93.96 & 94.88 & 94.81 \\ \cline{1-1} \cline{3-6}
        IterDet, 1 iter. && 94.34 & 94.27 & 94.97 & 94.89 \\
        IterDet, 2 iter. && \textbf{99.60} & \textbf{99.25} & \textbf{99.29} & \textbf{99.00} \\ \hline
    \end{tabular}
    \caption{Experimental results on AdaptIS Toy V1 and Toy V2 dataset.}
    \label{tab:toy}
\end{table*}

\begin{table*}[h!]
    \centering
    \begin{tabular}{c|c|ccc}
        \hline
        Method & Detector & Recall & AP & mMR  \\ \hline \hline
        Baseline \cite{shao2018crowdhuman} & \multirow{3}{*}{RetinaNet} & \textbf{93.80} & 80.83 & 63.33 \\ \cline{1-1} \cline{3-5}
        IterDet, 1 iter. && 79.68 & 76.78 & \textbf{53.03} \\
        IterDet, 2 iter. && 91.49 & \textbf{84.77} & 56.21 \\ \hline \hline
        Baseline \cite{shao2018crowdhuman} & \multirow{7}{*}{Faster RCNN} & 90.24 & 84.95 & 50.49 \\
        Soft NMS \cite{bodla2017soft,liu2019adaptive} && 91.73 & 83.92 & 51.97 \\
        Adaptive NMS \cite{liu2019adaptive} && 91.27 & 84.71 & 49.73 \\
        Repulsion Loss \cite{wang2017repulsion,ge2020ps} && 90.74 & 85.71 & - \\
        PS-RCNN \cite{ge2020ps} && 93.77 & 86.05& - \\ \cline{1-1} \cline{3-5}
        IterDet, 1 iter. && 88.94 & 84.43 & \textbf{49.12} \\
        IterDet, 2 iter. && \textbf{95.80} & \textbf{88.08} & 49.44 \\ \hline
    \end{tabular}
    \caption{Experimental results on CrowdHuman dataset with \emph{full-body} annotations.}
    \label{tab:crowd_human_full}
\end{table*}

\textbf{Results on WiderPerson.} The results on WiderPerson dataset are summarized in Table \ref{tab:wider_person}. We refer to \cite{zhang2019widerperson} for results obtained on \emph{hard} subset of annotations which contains all the boxes larger than 20 pixels in height. Following the protocol from \cite{ge2020ps}, we do not limit height during testing which is an even more challenging task. For both detectors, we achieve significantly better results in terms of recall, AP, and mMR.

\begin{table*}[h!]
    \centering
    \begin{tabular}{c|c|ccc}
        \hline
        Method & Detector & Recall & AP & mMR  \\ \hline \hline
        Baseline \cite{shao2018crowdhuman} & \multirow{4}{*}{RetinaNet} & \textbf{90.96} & 77.19 & 65.47 \\
        Feature NMS \cite{salscheider2020featurenms} && - & 68.65 & 75.35 \\ \cline{1-1} \cline{3-5}
        IterDet, 1 iter. && 86.91 & 81.24 & \textbf{58.78} \\
        IterDet, 2 iter. && 89.63 & \textbf{82.32} & 59.19 \\ \hline \hline
        Baseline \cite{shao2018crowdhuman} & \multirow{3}{*}{Faster RCNN} & 91.51 & \textbf{85.60} & 55.94 \\ \cline{1-1} \cline{3-5}
        IterDet, 1 iter. && 87.59 & 83.28 & \textbf{55.54} \\
        IterDet, 2 iter. && \textbf{91.63} & 85.33 & 55.61 \\ \hline
    \end{tabular}
    \caption{Experimental results on CrowdHuman dataset with \emph{visible-body}  annotations.}
    \label{tab:crowd_human_visible}
\end{table*}

\begin{table*}[h!]
    \centering
    \begin{tabular}{c|c|ccc}
        \hline
        Method & Detector & Recall & AP & mMR  \\ \hline \hline
        Baseline \cite{zhang2019widerperson} & \multirow{3}{*}{RetinaNet} & - & - & 48.32 \\ \cline{1-1} \cline{3-5}
        IterDet, 1 iter. && 90.38 & 87.17 & \textbf{43.23} \\
        IterDet, 2 iter. && \textbf{95.35} & \textbf{90.23} & 43.88 \\ \hline \hline
        Baseline \cite{zhang2019widerperson} & \multirow{5}{*}{Faster RCNN} & - & - & 46.06 \\
        Baseline \cite{ge2020ps} && 93.60 & 88.89 & - \\
        PS-RCNN \cite{ge2020ps} && 94.71 & 89.96 & - \\ \cline{1-1} \cline{3-5}
        IterDet, 1 iter. && 92.67 & 89.49 & \textbf{40.35} \\
        IterDet, 2 iter. && \textbf{97.15} & \textbf{91.95} & 40.78 \\ \hline
    \end{tabular}
    \caption{Experimental results on WiderPerson dataset.}
    \label{tab:wider_person}
\end{table*}

\begin{figure*}[t]
\centering
\setlength{\tabcolsep}{2pt}
\renewcommand{\arraystretch}{0.75}
\begin{tabular}{ccccc}
    \includegraphics[height=.862in]{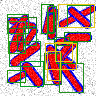} &
    \includegraphics[height=.862in]{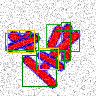} &
    \includegraphics[height=.862in]{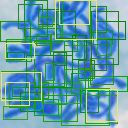} &
    \includegraphics[height=.862in]{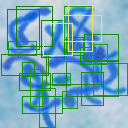} &
    \includegraphics[height=.862in]{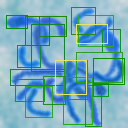} \\
\end{tabular}
\begin{tabular}{cc}
    \includegraphics[height=1.37in]{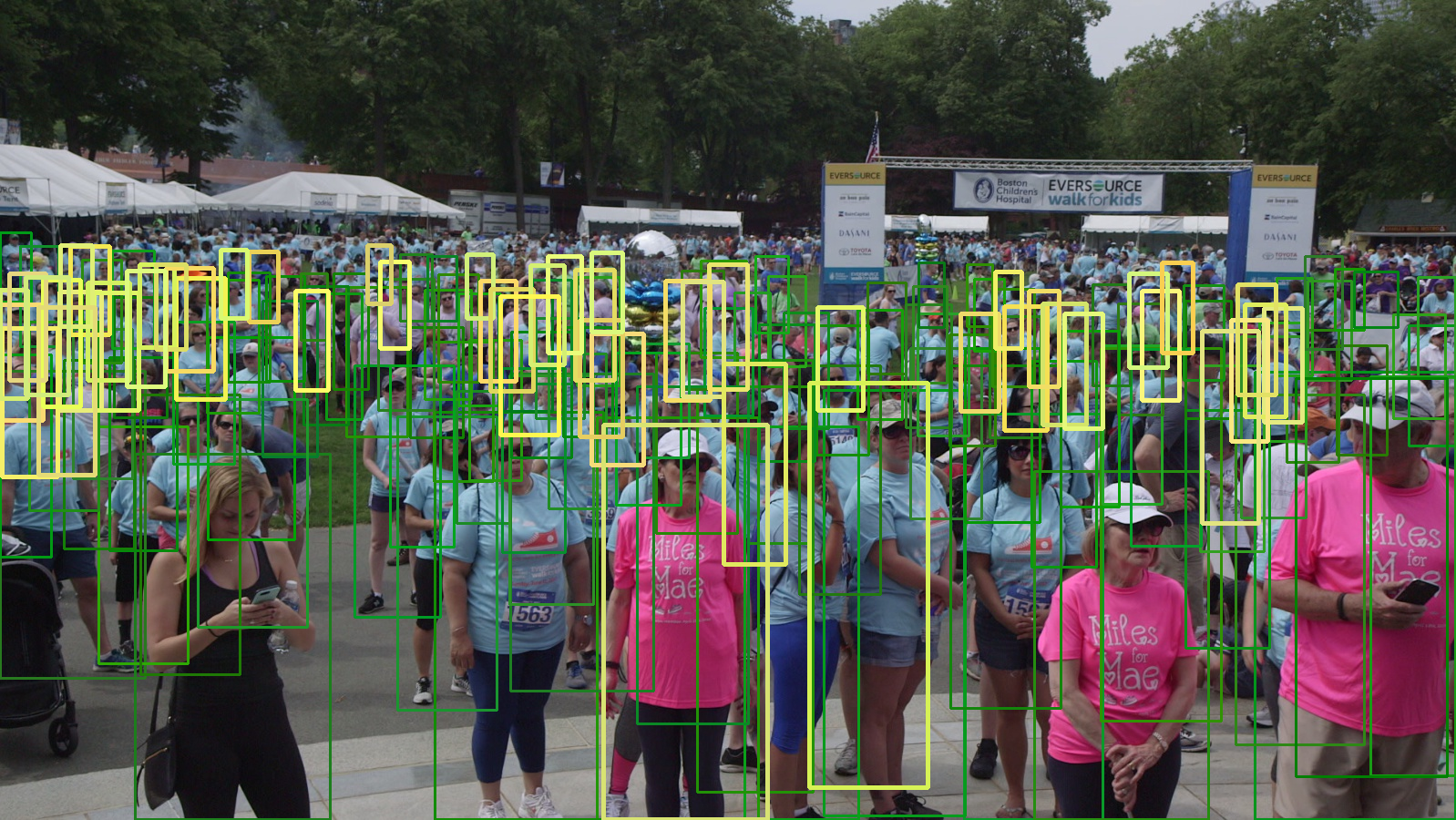} &
    \includegraphics[height=1.37in]{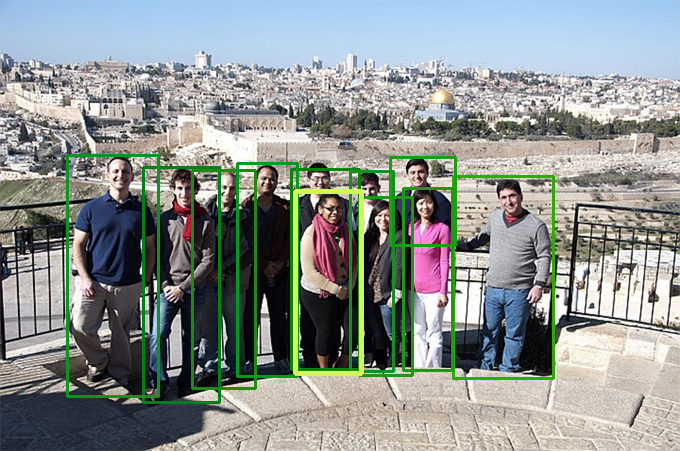} \\
\end{tabular}
\begin{tabular}{cc}
    \includegraphics[height=1.5in]{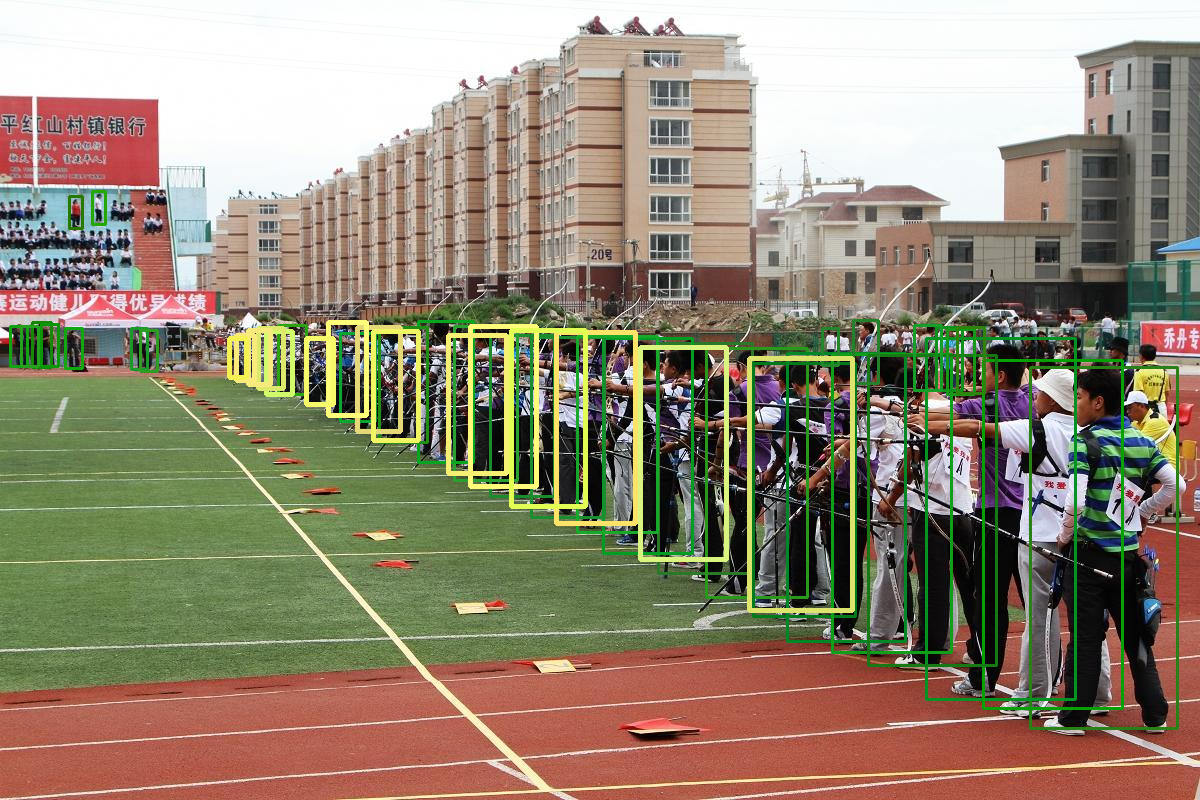} &
    \includegraphics[height=1.5in]{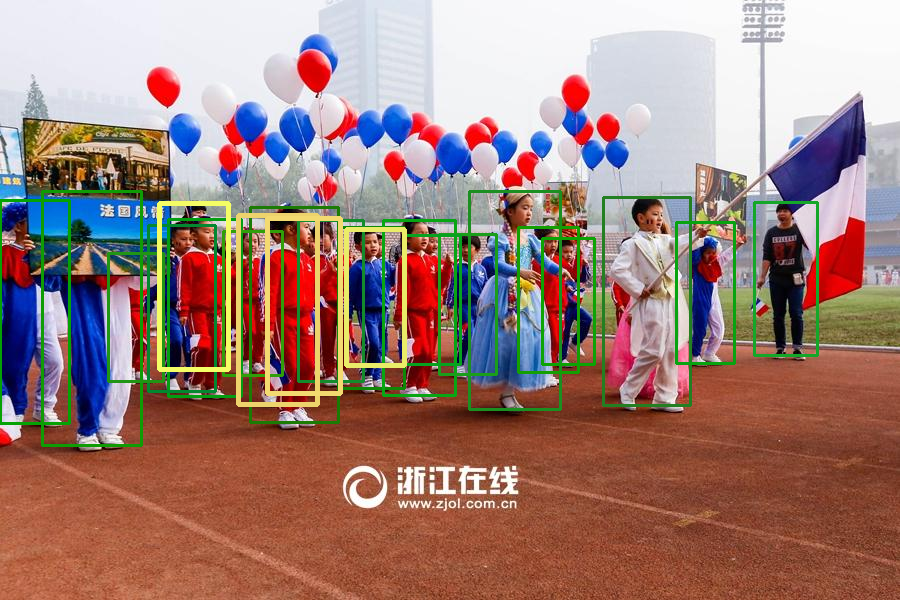} \\
\end{tabular}
\caption{IterDet results on ToyV1, ToyV2 (first row), CrowdHuman (with visible- and full-body annotations, second row), and WiderPerson (third row). The boxes found on the first and second iterations are marked in green and yellow respectively. The scores thresholded for visualization are above 0.1.}
\label{fig:examples}
\end{figure*}

Figure \ref{fig:examples} shows the results of IterDet based on Faster RCNN on the four datasets. In all examples, there are strongly overlapping objects with IoU$>$0.5 which are missed by the baseline detector but found by IterDet with 2 iterations.

\textbf{Choice of the number of iterations.} In some papers, e.g. \cite{ge2020ps}, only one metric out of AP and mMR is used for evaluation since their optimal values are not achieved simultaneously. In this work, we report the values of both metrics for CrowdHuman and WiderPerson datasets. Tables \ref{tab:crowd_human_full}, \ref{tab:crowd_human_visible} and \ref{tab:wider_person} demonstrate that the best value of mMR metric is achieved after the first iteration in the IterDet scheme. 

\begin{figure*}[h!]
    \begin{center}
        \includegraphics[width=0.6\linewidth]{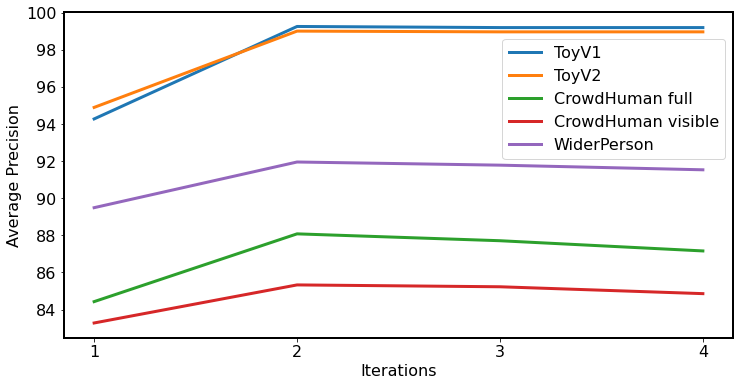}
    \caption{AP for different number of iterations for IterDet based on Faster RCNN.}
    \label{fig:ap}
    \end{center}
\end{figure*}
However, the optimal number of iterations in terms of AP metric is not as obvious. Figure \ref{fig:ap} depicts AP for a different number of iterations of the proposed iterative scheme based on Faster RCNN. For all datasets, we observe a noticeable improvement between the first and the second iteration. With increasing the number of iterations, AP does not improve. Moreover, for some datasets, a minor drop of AP can be observed. \\

We do not conduct experiments with a larger number of iterations due to the following reasons. First, IterDet already achieves state-of-the-art performance on CrowdHuman and WiderPersons datasets after only 2 iterations. Second, the inference time of the iterative scheme is proportional to the number of iterations, and for 3 iterations the inference would be 3 times slower which is not acceptable in practice.

\begin{figure*}[h!]
\centering
\setlength{\tabcolsep}{2pt}
\renewcommand{\arraystretch}{1.15}
\begin{tabular}{ccccccc}
    \includegraphics[height=0.65in]{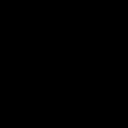} &
    \includegraphics[height=0.65in]{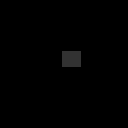} &
    \includegraphics[height=0.65in]{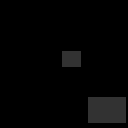} & &
    \includegraphics[height=0.65in]{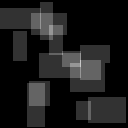} &
    \includegraphics[height=0.65in]{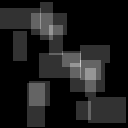} & \\
    \includegraphics[height=0.65in]{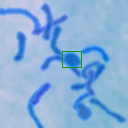} &
    \includegraphics[height=0.65in]{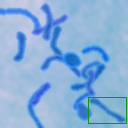} &
    \includegraphics[height=0.65in]{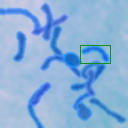} & &
    \includegraphics[height=0.65in]{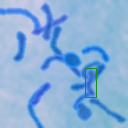} &
    \includegraphics[height=0.65in]{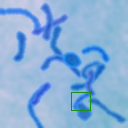} & 
    \multicolumn{1}{|c}{\includegraphics[height=0.65in]{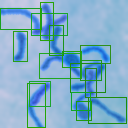}} \\
    1 & 2 & 3 & \ldots & 15 & 16 &
\end{tabular}
\caption{An additional experiment with limited detections per iteration for an image from Toy V2 test split. First row - history maps with already detected objects. Second row - an object detected on the corresponding iteration. Resulting detections are on the right.}
\label{fig:toy}
\end{figure*}

\textbf{Limiting detections per iteration.} We also provide the results of an additional experiment to prove the iterative scheme works. For this purpose, the only change is the restriction of one detection per iteration. This can be achieved by changing the NMS step to selection of the bounding box with the highest probability. Note that in this formulation the training process is not changed, and during inference the detector stops when 0 objects are predicted on the next iteration. The computational time of the detector on an image becomes proportional to the number of objects on it, which of course is not acceptable in practice. However, in term of metrics the proposed iterative scheme performs well. Thus, the AP on Toy V2 reaches 98.39\%, which is much higher than the baseline values from Table \ref{tab:toy}. The intermediate steps of the iterative scheme with limited detections per iteration are given on Figure \ref{fig:toy}. For a test image from Toy V2 with 16 objects all of them are successfully detected in 16 iterations.

\section{Conclusion}
\label{sec:conclusion}

We present an iterative scheme (IterDet) of object detection designed for crowded environments. It can be applied to both two-stage and one-stage object detectors. On challenging AdaptIS ToyV1 and ToyV2 datasets with multiple overlapping objects IterDet achieves almost perfect accuracy. Through extensive evaluation on CrowdHuman and WiderPerson benchmarks, we show that the proposed iterative scheme outperforms existing methods when applied to either two-stage Faster RCNN or one-stage RetinaNet detector.

{\small
\bibliographystyle{splncs04}
\bibliography{arxiv}
}

\end{document}